\documentclass[letterpaper, 9pt, twoside]{article}

\usepackage[english]{babel}

\usepackage[letterpaper,top=1.5cm,bottom=2cm,left=1.5cm,right=1.5cm,marginparwidth=1.75cm]{geometry}

\usepackage{amsmath}
\usepackage{float}
\usepackage{graphicx}
\usepackage{multicol}
\usepackage{enumitem}
\usepackage[colorlinks=true, allcolors=blue]{hyperref}
\usepackage{authblk}
\usepackage{booktabs}
\usepackage{caption}
\usepackage{fancyhdr}
\usepackage{placeins}  % For the \FloatBarrier command
\usepackage[textwidth=3.5cm]{todonotes}
\usepackage[backend=biber,style=nature,sorting=none]{biblatex}
\newcommand{\inserttitle}{
    Unsupervised risk factor identification across cancer types and data modalities via explainable artificial intelligence
}

\title{\textbf{\inserttitle}}

\author[1,2,3,4]{Maximilian Ferle$^{*\dagger}$}
\author[1,2]{Jonas Ader$^{\dagger}$}
\author[2,4]{Thomas Wiemers}
\author[2,4]{Nora Grieb}
\author[1,2]{Adrian Lindenmeyer}
\author[5]{Hans-Jonas Meyer}
\author[1,2]{Thomas Neumuth}
\author[3]{Markus Kreuz}
\author[1,3,6]{Kristin Reiche$^{\ddagger}$}
\author[4,7]{Maximilian Merz$^{\ddagger}$}

\affil[1]{Center for Scalable Data Analytics and Artificial Intelligence (ScaDS.AI) Dresden/Leipzig, Universität Leipzig, Germany}
\affil[2]{Innovation Center Computer Assisted Surgery (ICCAS), University of Leipzig, Leipzig, Germany}
\affil[3]{Department of Diagnostics, Fraunhofer Institute for Cell Therapy and Immunology, Leipzig, Germany}
\affil[4]{Department of Hematology, Hemostaseology, Cellular Therapy and Infectiology, University Hospital of Leipzig, Leipzig, Germany}
\affil[5]{Department of Diagnostic and Interventional Radiology, University Hospital Leipzig, Leipzig, Germany}
\affil[6]{Institute for Clinical Immunology, University Hospital of Leipzig, Leipzig, Germany}
\affil[7]{Memorial Sloan Kettering Cancer Center, New York, NY, USA}
\date{}
\usepackage{acro}

\DeclareAcronym{cnn}{
    short=CNN,
    long=Convolutional Neural Network,
}
\DeclareAcronym{mlp}{
    short=MLP,
    long=Multilayer Perceptron,
}
\DeclareAcronym{ct}{
    short=CT,
    long=Computed Tomography,
}
\DeclareAcronym{nsclc}{
    long=Non-Small Cell Lung Carcinoma,
    short=NSCLC,
}
\DeclareAcronym{shap}{
    short=SHAP,
    long=shapley additive explanation,
}
\DeclareAcronym{iccas}{
    short=ICCAS,
    long=Innovation Center Computer Assisted Surgery,
}
\DeclareAcronym{izi}{
    short=IZI,
    long=Institut für Zelltherapie und Immunologie,
}
\DeclareAcronym{dt}{
    short=DT,
    long=Digital Twin,
}
\DeclareAcronym{dmt}{
    short=DMT,
    long=Digital Medical Twin,
}
\DeclareAcronym{mm}{
    short=MM,
    long=Multiple Myeloma,
}
\DeclareAcronym{mmrf}{
    short=MMRF,
    long=Multiple Myeloma Research Foundation,
}
\DeclareAcronym{ml}{
    short=ML,
    long=machine learning,
}
\DeclareAcronym{riss}{
    short=R-ISS,
    long=Revised International Staging System,
}
\DeclareAcronym{mskcc}{
    short=MSKCC,
    long=Memorial Sloan Kettering Cancer Center,
}
\DeclareAcronym{hb}{
    short=Hb,
    long=Hemoglobin,
}
\DeclareAcronym{ca}{
    short=Ca,
    long=Calcium,
}
\DeclareAcronym{cr}{
    short=Cr,
    long=Creatinine,
}
\DeclareAcronym{wbc}{
    short=WBC,
    long=White blood bells,
}
\DeclareAcronym{mpr}{
    short=M-Pr,
    long=M-Protein,
}
\DeclareAcronym{sfl}{
    short=SFL,
    long=serum free light-chain
}
\DeclareAcronym{sflk}{
    short=SFL-$\kappa$,
    long=serum free light-chain $\kappa$
}
\DeclareAcronym{sfll}{
    short=SFL-$\lambda$,
    long=serum free light-chain $\lambda$
}
\DeclareAcronym{ldh}{
    short=LDH,
    long=Lactate dehydrogenase,
}
\DeclareAcronym{alb}{
    short=Alb,
    long=Albumin,
}
\DeclareAcronym{b2m}{
    short=$\beta$2m,
    long=$\beta$-2-microglobulin,
}
\DeclareAcronym{tnm}{
    short=TNM,
    long=TNM Classification of Malignant Tumors,
}
\DeclareAcronym{zpce}{
    short=ZPCE,
    long=zero-signal preserving contrast enhancement,
}
\DeclareAcronym{ila}{
    short=ILA,
    long=interstitial lung abnormalities,
}
\DeclareAcronym{prc}{
    short=PRC,
    long=precision-recall curve,
}

\begin{document}
    \maketitle
    \def\thefootnote{*}\footnotetext{Correspondence to: maximilian.ferle@uni-leipzig.de}\def\thefootnote{\arabic{footnote}}
    \def\thefootnote{$\dagger$}\footnotetext{These authors contributed equally and share first authorship.}\def\thefootnote{\arabic{footnote}}
    \def\thefootnote{$\ddagger$}\footnotetext{These authors contributed equally and share last authorship.}\def\thefootnote{\arabic{footnote}}

    \begin{abstract}
        Risk stratification is a key tool in clinical decision-making, yet current approaches often fail to translate sophisticated survival analysis into actionable clinical criteria.
We present a novel method for unsupervised machine learning that directly optimizes for survival heterogeneity across patient clusters through a differentiable adaptation of the multivariate logrank statistic.
Unlike most existing methods that rely on proxy metrics, our approach represents a novel method for training any neural network architecture on any data modality to identify prognostically distinct patient groups.
We thoroughly evaluate the method in simulation experiments and demonstrate its utility in practice by applying it to two distinct cancer types: analyzing laboratory parameters from multiple myeloma patients and computed tomography images from non-small cell lung cancer patients, identifying prognostically distinct patient subgroups with significantly different survival outcomes in both cases.
Post-hoc explainability analyses uncover clinically meaningful features determining the group assignments which align well with established risk factors and thus lend strong weight to the methods utility.
This pan-cancer, model-agnostic approach represents a valuable advancement in clinical risk stratification, enabling the discovery of novel prognostic signatures across diverse data types while providing interpretable results that promise to complement treatment personalization and clinical decision-making in oncology and beyond.
    \end{abstract}

    \begin{figure}[H]
        \includegraphics[width=\linewidth]{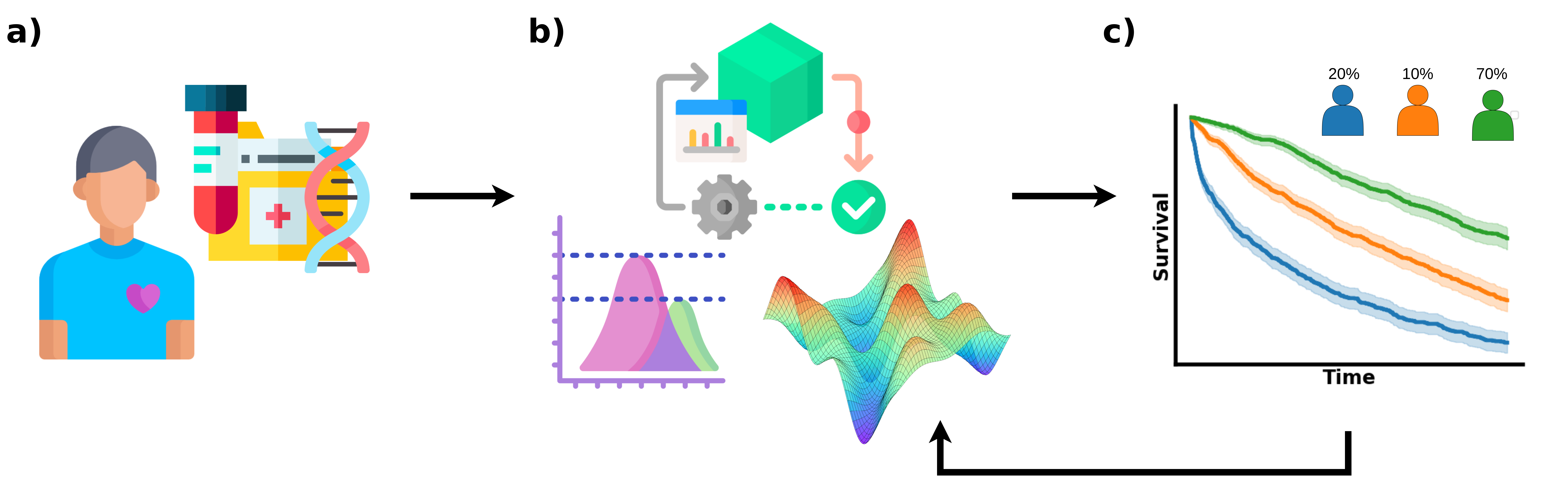}
        \caption{
            Graphical abstract of the machine learning pipeline for patient survival clustering.
            The workflow progresses from raw patient data collection through algorithmic processing to final risk classification, enabling personalized prognostic assessment based on multivariate clinical parameters.
            (\textbf{a}) Input data comprising multimodal clinical patient data such as laboratory parameters and genetic information.
            (\textbf{b}) Development phase showing the implementation of a custom machine learning algorithm.
            The panel includes, algorithm optimization (top), model training (lower right), and validation steps (lower left).
            (\textbf{c}) Visualization of the desired model output displaying survival outcomes of patient risk clusters and partial attribution of patients to each group.
            The graphical abstract was designed with the open source software \textit{draw.io} using resources from \textit{Flaticon.com}.
        }
        \label{fig:graphical_abstract}
    \end{figure}

    \begin{multicols}{2}

        \section{Introduction}\label{sec:introduction}
        Survival analysis is a vital statistical methodology in healthcare that examines the time until clinical events such as disease progression or death occur while accounting for variable follow-up periods.
Risk stratification, built on these principles, guides treatment decisions and resource allocation across a variety of domains by enabling clinicians to categorize patients into groups of differing prognosis based on prognostic markers\supercite{clark_survival_2003}.
While statistical approaches like Cox proportional hazards models\supercite{harrell_cox_2001} and random survival forests\supercite{ishwaran_random_2008} enable researchers to quantify the relative impact of variables on survival outcomes, these methods face operational challenges in clinical practice, because they often fall short of providing clear decision boundaries\supercite{eng_representing_2015, wang_comparison_2023} that would facilitate straightforward stratification into risk groups and their interpretation requires expert domain knowledge\supercite{sarica_explainability_2023, innes_competing_2022}.
Due to this limitation, these models find little direct application in day-to-day clinical decision-making\supercite{hu_personalized_2018, ahmed_systematic_2023-1}.
One example of clinically useful patient grouping is demonstrated by the \ac{riss}\supercite{palumbo_revised_2015} for \ac{mm}, which has proven its practical value in oncology care by effectively stratifying patients into three distinct risk groups based on clearly defined boundaries for laboratory parameters and cytogenetic findings.
The success of the \ac{riss} in clinical practice stems from its ability to translate complex prognostic information into clear, actionable decision points that guide treatment selection and intensity\supercite{jimenez-zepeda_revised_2016}.
While sophisticated machine learning approaches have revolutionized many aspects of survival analysis\supercite{wiegrebe_deep_2024}, their potential to aid patient grouping based on risk factors remains surprisingly limited in scope.
Current methodologies mostly either rely on clustering patients based on feature similarity metrics that may not necessarily correlate with survival patterns\supercite{chapfuwa_survival_2020, manduchi_deep_2022}, employing recursive partitioning techniques that offer little insight on an individual patient basis\supercite{you_survivalclusteringtree_2023}, or modeling survival functions through elaborate mixture models that may not capture the influence of underlying biological predictors\supercite{jeanselme_neural_2022, buginga_clustering_2024} or require a priori knowledge of survival times\supercite{mouli_deep_2019}.
The challenge of clustering survival data based on patient features has remained largely unaddressed in computational medicine, with existing approaches typically focusing on proxy metrics, but not their direct relationship.
While recent efforts have proposed dedicated model architectures for specific data modalities\supercite{ahmad_towards_2017, qiu_deep_2025, de_jong_deep_2019}, the problem of survival-based clustering lacks a unified optimization approach that generalizes across different model architectures and the full spectrum of clinical data modalities.
Our work addresses this gap by shifting focus from model architecture to the parameterization of the optimization problem itself.
By proposing an optimization criterion derived from the multivariate logrank statistic, we provide a flexible framework that enables any neural network architecture to learn survival-based clustering directly from patient features in an unsupervised manner.
This architecture-agnostic approach represents a paradigm shift in machine learning methodology, as demonstrated by our successful application to both \ac{mlp}s processing laboratory parameters and \ac{cnn}s analyzing \ac{ct} images, identifying prognostically distinct patient subgroups with significantly different survival outcomes in both cases.

        \section{Results}\label{sec:results}
        \subsection{Partial Multivariate Logrank Loss}\label{sec:optimization_criterion}
We developed a novel approach to patient stratification by reformulating the multivariate logrank statistic as a differentiable optimization criterion suitable for neural network training.
Our implementation leverages PyTorch's\supercite{paszke_pytorch_2019} automatic differentiation capabilities through a custom loss module, which we coined \textit{PartialMultivariateLogRankLoss}.
The framework leverages predictions of probability distributions $p_{i,g}$ over possible group assignments, while the classical logrank statistic uses binary indicators $\delta_{g,i}$ (refer to methods section \textit{\nameref{sec:eqns}}).
This generalization enables backpropagation through the computational graph, while maintaining mathematical consistency with the original statistic.
The module processes three essential inputs: predicted group assignment probabilities, survival times, and event indicators to compute the \textit{PartialMultivariateLogRankLoss} as a measure for survival heterogeneity as the optimization criterion.
Rather than developing specialized model architectures, we thereby formalized the core optimization problem of survival-based clustering itself, creating a universal framework that enables any neural network to stratify patients irrespective of data modality.

During early development, we encountered challenges related to trivial solutions in the loss calculation.
As the logrank is a convex function (imbalanced group assignments maximize the logrank), we augmented the loss function with a penalty term to promote balanced group assignments:
\begin{equation}
    L_\text{total} = L_\text{logrank} - \lambda P(p)
\end{equation}\label{eq:total_loss}
where $P(p)$ penalizes imbalanced group assignments with penalty weight $\lambda$, which can be adjusted through hyperparameter tuning.
The penalty function is designed to encourage uniform group sizes through a transformation that maintains differentiability:
\begin{equation}
    P(p) = \frac{1}{k} \sum_{i=1}^{k} \frac{1}{p_i^\alpha - (p_i^\alpha)^2} - 4
\end{equation}
with $\alpha = \frac{\ln(1/2)}{\ln(1/k)}$ for $k$ groups.
This formulation creates an asymmetric barrier that bottoms out at $1/k$, effectively counteracting the tendency of the logrank loss to collapse to trivial solutions.

To validate our proposed framework, we implemented a structured experimental approach to test our methodology across diverse data modalities and neural network architectures.

First, we conducted a simulation experiment with synthetic tabular data where we established ground truth groupings and systematically evaluated a \ac{mlp}'s ability to recover these predefined patient clusters.
Second, we applied a \ac{cnn} in controlled simulation experiments to cluster handwritten digits based on associated simulated survival times, establishing ground truth recovery performance in an optical recognition setting.
These validation approaches were necessitated by the unsupervised nature of our method, which precludes knowledge of true groupings in real-world clinical data.

Building on these simulation studies, we next sought to evaluate the generalizability of our approach by applying it to real-world clinical datasets from two fundamentally distinct tumor entities | one hematologic and one solid malignancy | each characterized by entirely different input modalities: laboratory parameters in the former and imaging-derived features (\ac{ct} scans) in the latter.
To this end, we employed a \ac{mlp} to cluster \ac{mm} patients based on laboratory blood parameters, evaluating its performance in separating patients into prognostically distinct risk groups based on progression-free survival.
We conducted explainability analyses to identify which laboratory parameters most strongly influenced the model's clustering decisions to gain insight into the prognostic features leveraged for patient stratification.
Finally, we advanced to cluster \ac{nsclc} patients based on \ac{ct} imaging data using a \ac{cnn}, achieving significant separation in survival outcomes.
We further employed visualization techniques to identify attention patterns within the images that drove the model's decisions, ultimately revealing morphological features associated with prognosis that align with conventional staging approaches.

The results of our analyses are detailed below.

\subsection{Performance assessment of our method using a \ac{mlp} on simulated tabular data}
\begin{figure*}[ht!]
    \centering
    \includegraphics[width=0.8\linewidth]{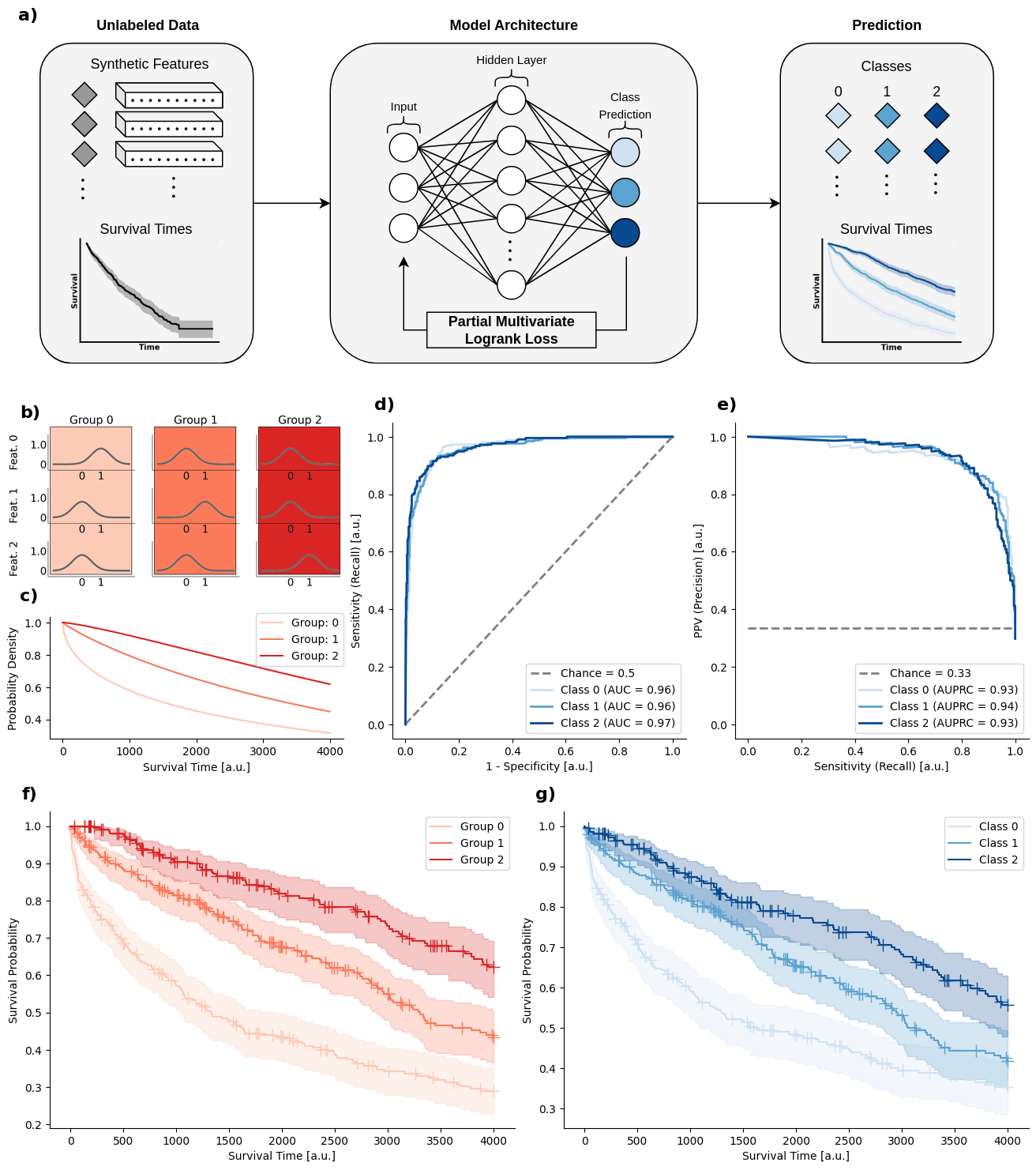}
    \caption{
        \textbf{Stratification of synthetic feature vectors according to their associated survival distributions.}
        \textbf{(a)} Architecture diagram showing the multilayer perceptron workflow: synthetic feature vectors with survival times (left) are clustered through the network using our custom \textit{PartialMultivariateLogrankloss} (center) to generate class predictions (right).
        \textbf{(b)} Probability density of the multivariate gaussian distributions used to sample the synthetic feature vectors from, color-coded by the three ground truth groups.
        \textbf{(c)} Weibull distributions used to sample survival times for each group from.
        \textbf{(d)} ROC curves showing performance of ground truth recovery with AUROC values of 0.96, 0.96, and 0.97 for classes 0, 1, and 2, respectively.
        \textbf{(e)} Precision-recall curves (PRC) showing performance of ground truth recovery with AUPRC values of 0.93, 0.94, and 0.93 for classes 0, 1, and 2, respectively.
        \textbf{(f)} Kaplan-Meier survival curves with 95\% confidence intervals for the resulting three ground truth groups.
        \textbf{(g)} Kaplan-Meier survival curves with 95\% confidence intervals for the three classes identified by the multilayer perceptron without prior knowledge of ground truth groupings.
    }
    \label{fig:mlp_fig}
\end{figure*}

To validate our method in the common clinical scenario of analyzing structured patient data (Figure \ref{fig:mlp_fig}a), we generated synthetic datasets comprising three-dimensional feature vectors, where each dimension represented a continuous variable analogous to laboratory parameters or clinical measurements commonly encountered in medical practice.

This controlled simulation experiment addresses the challenge of survival-based clustering without knowledge of a definitive ground truth, a problem inherent to unsupervised approaches.
By creating a synthetic ground truth association between specific feature distributions and survival outcomes, we could systematically evaluate our method's ability to recover these predefined relationships.

We hence defined three distinct ground truth groups by sampling feature vectors from overlapping multivariate normal distributions (Figure \ref{fig:mlp_fig}b).
This approach models the inherent biological variability and partial overlap of biomarker profiles observed in clinical populations, where patient subgroups rarely exhibit completely distinct feature distributions.
The partial overlap between distributions creates a challenging yet realistic scenario for testing our clustering algorithm's discriminative capabilities.

For each ground truth group, we sampled survival times from group-specific Weibull distributions (Figure \ref{fig:mlp_fig}c).
We selected Weibull distributions as it is a common choice in modeling survival distributions due to their flexibility in modeling various hazard patterns commonly observed in clinical survival data\supercite{plana_cancer_2022, majer_estimating_2022}.
The three distributions were parameterized to represent distinct risk profiles based in real-world evidence (refer to methods section \textit{\nameref{sec:surv_sim}} for details).
Additionally, to create the characteristic incomplete follow-up pattern common in survival analysis, we introduced random right-censoring by generating competing censoring times from an exponential distribution.
The Kaplan-Meier curves resulting from sampling survival times for these ground truth groups (Figure \ref{fig:mlp_fig}d) demonstrate clear separation.

We hence implemented a \ac{mlp} architecture (Figure \ref{fig:mlp_fig}a; refer to methods section \textit{\nameref{sec:model_training}} for details) to map the three-dimensional feature vectors to group assignments.
Critically, the \ac{mlp} was trained exclusively using our custom \textit{PartialMultivariateLogrankLoss} function, without any knowledge of the predefined group assignments and learned survival-relevant patterns solely through the optimization of between-group survival heterogeneity.

The \ac{mlp} demonstrated robust performance in recovering the underlying survival patterns from the synthetic tabular data.
Quantitative assessment through ROC analysis (Figure \ref{fig:mlp_fig}d) confirmed the model's strong discriminative performance, achieving AUROC values of 0.96, 0.96, and 0.97 for Classes 0, 1, and 2, respectively.
The \ac{prc} analysis (Figure \ref{fig:mlp_fig}e) also showed strong performance, with AUPRC values of 0.93, 0.94, and 0.93 for Classes 0, 1, and 2, respectively, indicating high precision of the models predictions across all three classes.
These values indicate that by training the model to optimize between-group survival heterogeneity using our custom \textit{MultivariateLogrankLoss}, the \ac{mlp} can effectively distinguish between the three risk groups despite the challenging scenario of overlapping feature distributions and censored survival data.

Comparison of the ground truth Kaplan-Meier survival curves (Figure \ref{fig:mlp_fig}f) with those identified by the model (Figure \ref{fig:mlp_fig}g) reveals excellent concordance, with the model successfully reproducing the distinct survival trajectories of each group.
\subsection{Performance assessment of our method using a \ac{cnn} on image data}
\begin{figure*}[ht!]
    \centering
    \includegraphics[width=0.8\linewidth]{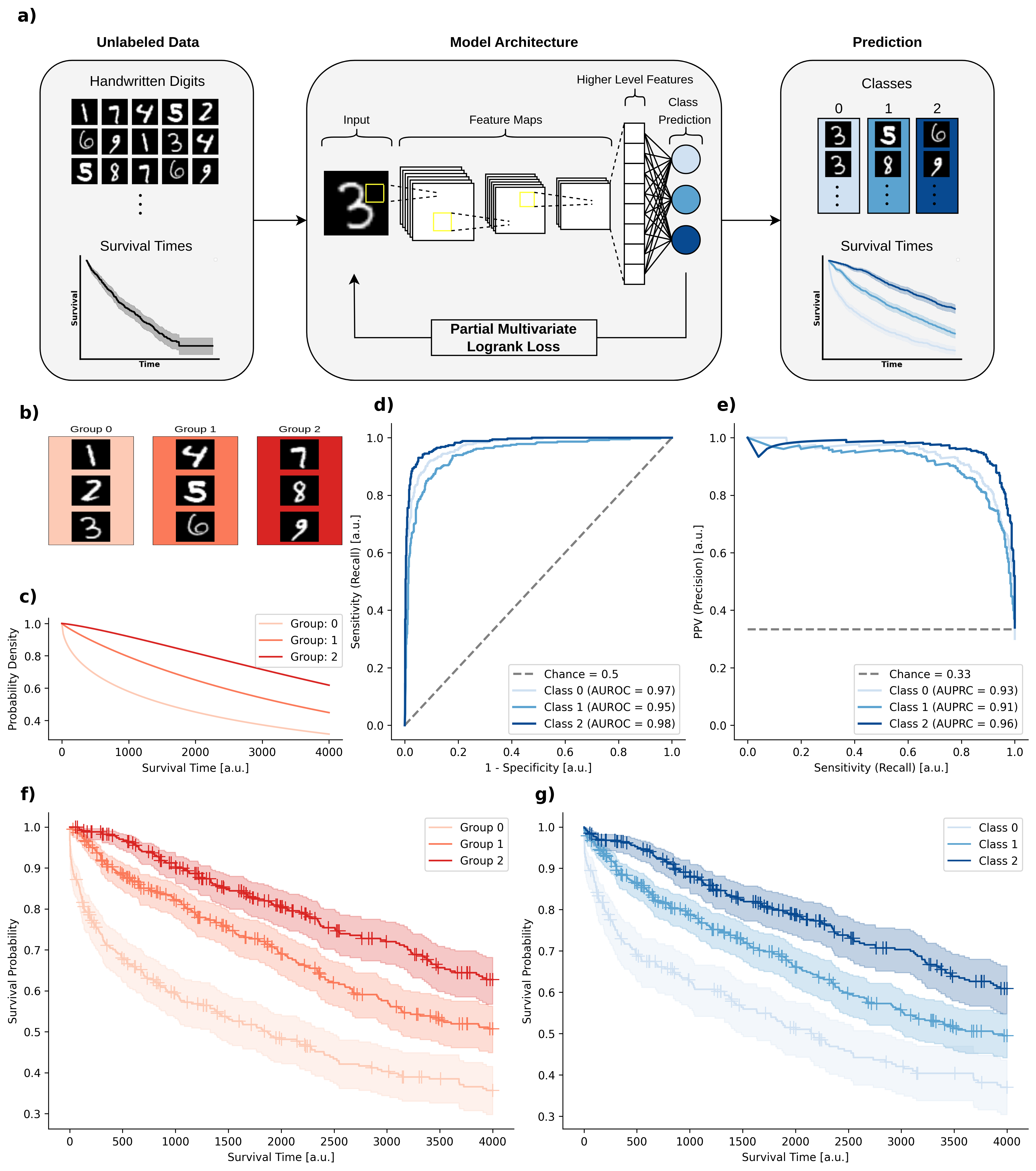}
    \caption{
        \textbf{Stratification of handwritten digits according to their associated survival distributions.}
        \textbf{(a)} Architecture diagram showing the \ac{cnn} workflow: MNIST handwritten digits with associated survival times (left) are processed through convolutional layers to extract features (center) and generate class predictions (right) using our custom \textit{PartialMultivariateLogrankLoss}.
        \textbf{(b)} Representative MNIST digits from the three predefined groups: Group 0 (light red) contains digits 1-3, Group 1 (medium red) includes digits 4-6, and Group 2 (dark red) comprises digits 7-9.
        \textbf{(c)} Weibull distributions used to sample survival times for each digit group.
        \textbf{(d)} ROC curves showing discriminative performance of ground truth groupings with AUROC values of 0.97, 0.95, and 0.98 for Classes 0, 1, and 2, respectively.
        \textbf{(e)} Precision-recall curves (PRC) showing precision of models' predictions with AUPRC values of 0.93 0.91, and 0.96 for Classes 0, 1, and 2, respectively.
        \textbf{(f)} Kaplan-Meier survival curves with 95\% confidence intervals for the resulting three ground truth groups.
        \textbf{(g)} Kaplan-Meier curves for the three classes identified by the \ac{cnn} without prior knowledge of ground truth groupings.
    }
    \label{fig:mnist_fig}
\end{figure*}

To further demonstrate the versatility of our approach across different data modalities and model architectures, we extended our validation to the domain of computer vision.
To simulate the clinical setting of pattern recognition in heterogeneous imaging data, we used the MNIST dataset of handwritten digits\supercite{e_alpaydin_optical_1998, li_deng_mnist_2012}.
While maintaining the same core optimization framework based on our \textit{PartialMultivariateLogrankLoss}, we transitioned from the \ac{mlp} architecture to a \ac{cnn} suited for image processing.
Again, we used a controlled simulation experiment to evaluate the performance of our method to overcome the absence of ground truth for optimal survival-based clustering in real-world imaging data, which is inherent to the unsupervised nature of our approach.

Here, we parameterized our experiment analogous to before by associating specific handwritten digits with distinct survival distributions.
To this end, the digits were organized into three predefined groups (Figure \ref{fig:mnist_fig}b) while, for each digit, we simulated survival times by sampling from a group-specific Weibull distributions (Figure \ref{fig:mnist_fig}c) accounting for incomplete follow-ups analogous to our previous experiment.
By doing so, we could create a clear association between visual patterns and survival outcomes, which simulated the clinical situation of imaging data being associated with different outcomes, and measure our method's ability to recover these predefined relationships.

We implemented a \ac{cnn} (Figure \ref{fig:mnist_fig}a) to process the digit images and map them to group assignments.
Critically, as with our previous \ac{mlp} experiment, the \ac{cnn} was trained exclusively using our \textit{PartialMultivariateLogrankLoss} function, without any knowledge of the predefined group assignments or digit identities.
The model learned to identify survival-relevant patterns solely through optimization of between-group survival heterogeneity.

The \ac{cnn} demonstrated excellent performance in recovering the underlying survival patterns from the image data.
Quantitative assessment through ROC analysis (Figure \ref{fig:mnist_fig}d) confirmed the model's strong discriminative performance, achieving AUC values of 0.97, 0.95, and 0.98 for Classes 0, 1, and 2, respectively.
The \ac{prc} analysis (Figure \ref{fig:mlp_fig}e) also showed strong performance, with AUPRC values of 0.93, 0.91, and 0.96 for Classes 0, 1, and 2, respectively, indicating high precision of the models' predictions across all three classes.
These values indicate that our method enabled the \ac{cnn} to effectively distinguish between the three groups solely based on the survival times associated with the image data.
Comparison of the ground truth Kaplan-Meier survival curves (Figure \ref{fig:mnist_fig}f) with those identified by the model (Figure \ref{fig:mnist_fig}g) reveals remarkable concordance, with the model successfully reproducing the distinct survival trajectories of each group.

The successful application of our methodology to this computer vision task, using the same core optimization approach highlights its robustness across a different network architectures and data modalities.
This adaptability suggests promising applications in diverse clinical settings where different data types can be leveraged to identify prognostically distinct patient groups across various cancer types and diverse clinical data modalities.

Encouraged by these strong results in controlled settings, we extended our analyses to applying our framework to a set of challenging real-world clinical problems.

\subsection{Unsupervised risk stratification of \ac{mm} patients using routine blood work}\label{sec:myeloma-mlp}
\begin{figure*}[ht!]
    \includegraphics[width=\linewidth]{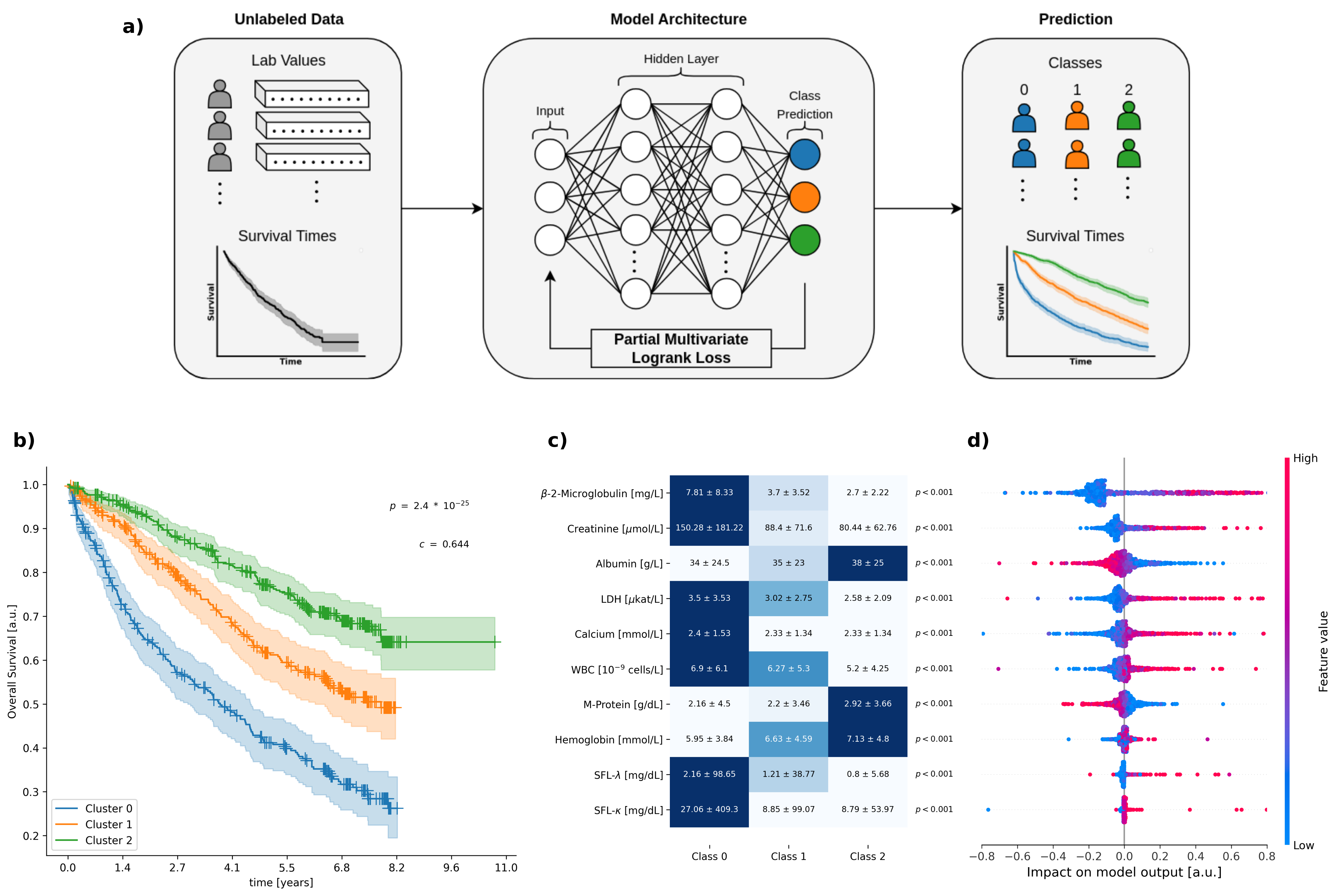}
    \caption{
        \textbf{Application of our method by stratifying \ac{mm} patients based on biomarker profiles.}
        \textbf{a)} Workflow demonstrating unlabeled patient data (left) encompassing lab values and survival times processed by a \ac{mlp} (center) trained on our custom \textit{PartialMultivariateLogrankLoss} to generate class predictions (right) categorized into three classes (0, 1, 2) with corresponding survival time curves.
        \textbf{b)} Overall survival analysis showing significant stratification of patients into three distinct risk clusters ($p = 2.4 \times 10^{-25}$, $c = 0.644$).
        Cluster 0 (blue) demonstrates the poorest outcomes with median survival at approximately at 4 years, Cluster 1 (orange) shows intermediate outcomes, while Cluster 2 (green) shows the most favorable prognosis with approximately 70\% survival rate at 9 years.
        \textbf{c)} Biomarker distribution (mean ± s.d.) across the three patient classes for various laboratory parameters including $\beta$-2-Microglobulin, Creatinine, M-Protein, Calcium, Albumin, LDH, WBC, Hemoglobin, SFL-$\kappa$, and SFL-$\lambda$.
        Highly significant differences ($p < 0.001$) were observed for all parameters based on a Kruskal-Wallis test across the three classes.
        Color gradients (light to dark) visually represent feature magnitudes (low to high respectively) normalized by each parameter's range indivudally.
        \textbf{d)} Feature importance analysis showing the impact of key biomarkers on model output, with beeswarm plots displaying SHAP values for the biomarkers shown in panel c at the same height.
        The visualization uses a color gradient (blue to red) to indicate feature values from low to high, with dots showing the distribution of impact values across the patient population.
        All shown data were derived from the testing partitions of the 5 cross-validation-folds and are entirely based on unseen patients.
    }
    \label{fig:blood_cluster_fig}
\end{figure*}

To demonstrate the clinical utility of our optimization framework, we applied it to stratify \ac{mm} patients based solely on routine blood work parameters (Figure \ref{fig:blood_cluster_fig}a).
We utilized the CoMMpass (Clinical Outcomes in \ac{mm} to Personal Assessment) dataset, which represents one of the largest publicly available clinical datasets in \ac{mm} research.
This dataset is particularly suitable for validating our framework as it contains comprehensive biomarker profiles alongside detailed clinical outcomes, enabling robust evaluation of risk stratification approaches.

Furthermore, \ac{mm} was selected as the real-world use case due to the availability of well-established risk stratification systems developed over the past decades, including the ISS\supercite{greipp_international_2005}, the \ac{riss}\supercite{palumbo_revised_2015}, and more recently, the R2-ISS\supercite{dagostino_second_2022}.
These staging systems provide clinically meaningful survival prognostications based on specific biomarkers.
The inclusion of \ac{mm} thus enables us to benchmark our method's ability to recover established, biologically relevant prognostic features | such as \ac{b2m}, \ac{alb}, and others.

Based on these considerations, we selected ten key biomarkers routinely collected in clinical practice and known to be relevant in \ac{mm}: \ac{hb}, \ac{ca}, \ac{cr}, \ac{ldh}, \ac{alb}, \ac{b2m}, \ac{mpr}, \ac{sfll}, \ac{sflk}, and \ac{wbc} (refer to methods section \textit{\nameref{sec:patient_characteristics}} for detailed rationale).
A \ac{mlp} model was trained to cluster patients into three distinct risk groups based on these parameters (Figure \ref{fig:blood_cluster_fig}a).
The selection of three clusters was motivated by the beforementioned established clinical risk stratification systems in \ac{mm}, such as the \ac{riss}\supercite{palumbo_revised_2015}.

In accordance to best practices in machine learning, we partitioned the dataset using 5-fold cross-validation and would exclusively evaluate model performance in unseen patients belonging to the withheld testing partitions (Figure \ref{fig:blood_cluster_fig}b-d).
The model achieved a highly significant separation of the overall survival ($p = 2.4 \times 10^{-25}$, $c = 0.644$, Figure \ref{fig:blood_cluster_fig}b), with distinct survival trajectories across the three clusters.
Cluster 0 exhibited the poorest prognosis with median survival of approximately 4 years, Cluster 1 demonstrated intermediate outcomes, while Cluster 2 showed markedly better prognosis with approximately 70\% survival at 9 years.
Notably, our model achieved a concordance index (c-index) of 0.644, which is of a similar magnitude to the typically reported c-index for the \ac{riss} in \ac{mm} in different cohorts, which typically are on the order of 0.61\supercite{cho_evaluation_2024, brieghel_real-world_2025} and the recently proposed IrMMa system\supercite{maura_genomic_2024}, which reports a c-index of 0.687.
This performance is particularly significant considering that the \ac{riss} relies on cytogenetic data and the IrMMa system relies on genomic data, which both are normally not available in routine clinical settings, whereas our approach utilizes only blood parameters that accumulate naturally during standard clinical care.

Analysis of biomarker distributions across the three identified clusters revealed clinically meaningful patterns that align with established disease biology (Figure \ref{fig:blood_cluster_fig}c).
The high-risk Cluster 0 exhibited significantly elevated levels of known adverse prognostic markers, including \ac{b2m} ($p~<~0.001$), \ac{cr} ($p~<~0.001$), \ac{ldh} ($p~<~0.001$), \ac{ca} ($p~<~0.001$), \ac{wbc} ($p~<~0.001$) and \ac{sfl} ($p~<~0.001$), alongside reduced \ac{alb} ($p~<~0.001$), \ac{mpr} ($p~<~0.001$) and \ac{hb} ($p~<~0.001$) levels.
Conversely, the favorable-risk Cluster 2 demonstrated lower \ac{b2m} and higher \ac{alb} levels, consistent with established biomarkers indicating less advanced disease.
All biomarkers showed statistically significant differences across the three clusters based on a Kruskal-Wallis test, confirming the biological relevance of the identified patient subgroups.

Feature importance analysis using \ac{shap} values (Figure \ref{fig:blood_cluster_fig}d) provided further insights into the model's decision-making process.
\ac{shap} values are an estimation of classic Shapley values from game theory\supercite{kuhn_17_1953}, which in the context of machine learning allows to approximate each feature's contribution to a given model's predictions\supercite{ferle_predicting_2025}.
\ac{b2m} and \ac{cr} emerged as the most influential parameters, with higher values (red dots) being associated with higher risk, consistent with their respective roles in renal dysfunction and disease progression in \ac{mm}.
 Elevated \ac{alb} and \ac{hb} showed association with lower risk and improved outcomes, reflecting the adverse impact of systemic inflammation or impaired protein synthesis and anemia on patient prognosis.
These patterns strongly align with clinical expectations and established disease biology, providing compelling validation of our method to uncover biologically relevant risk patterns.

Notably, our unsupervised approach identified these clinically relevant patient subgroups without any prior knowledge of established risk factors or staging systems.
The model learned to recognize complex patterns in the biomarker data solely through optimization of survival heterogeneity between the identified clusters.

These findings highlight the utility of our unsupervised survival-based clustering framework for patient stratification in \ac{mm} and suggest potential applications in other malignancies where prognostic biomarkers are routinely collected but optimal risk stratification remains challenging.
\subsection{Uncovering prognostic signatures in \ac{ct} images of \ac{nsclc} patients}

\begin{figure*}[th!]
    \centering
    \includegraphics[width=\linewidth]{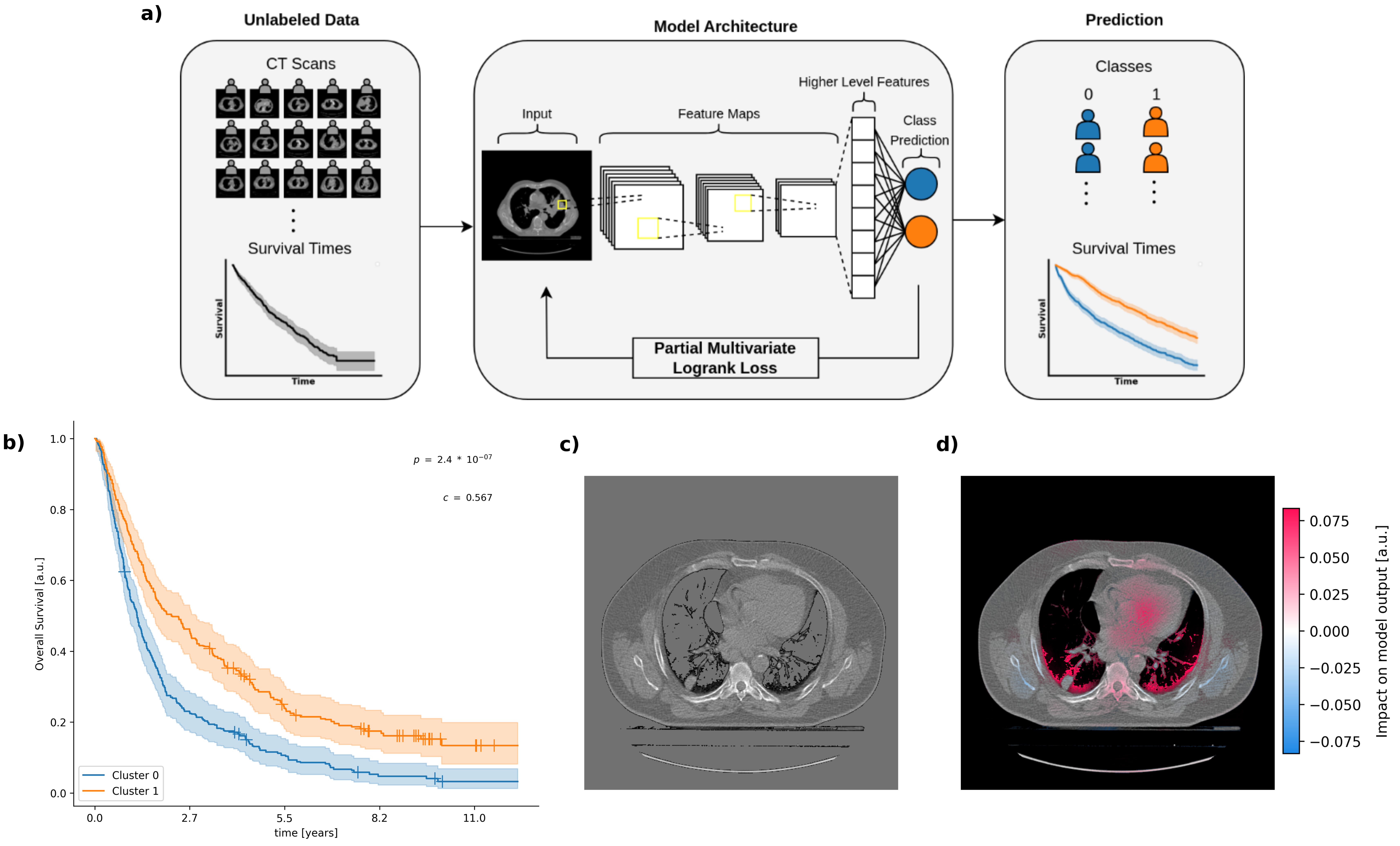}
    \caption{
        \textbf{Application of our method by stratifying \ac{nsclc} patients based on \ac{ct} imaging data.}
        \textbf{a)} Radiomics workflow demonstrating the application of a \ac{cnn} to \ac{nsclc} \ac{ct} images to classify patients into distinct risk groups (high vs.
        low) based on our custom \text{PartialMultivariateLogrankLoss}.
        \textbf{b)} Kaplan-Meier survival curves showing significant difference ($p=2.4\times10^{-7}$) between the two patient clusters identified by the radiomics model.
        Cluster 0 (blue) exhibits poorer survival outcomes compared to Cluster 1 (orange).
        \textbf{c)} Representative \ac{ct} image with zero-signal preserving contrast enhancement.
        \textbf{d)} \ac{shap} value heatmap corresponding to \ac{ct} image in panel c, illustrating pixel-level importance for the model's survival-based predictions.
        Color scale bars indicate \ac{shap} value magnitude (arbitrary units), representing increased model attention in colored areas.
        Red areas indicate features positively associated with high-risk classification, while blue areas represent features associated with lower risk.
        All shown data were derived from the testing partitions of the 5 cross-validation-folds and are entirely based on unseen patients.
    }
    \label{fig:lung_cluster_fig}
\end{figure*}

\begin{figure*}[th!]
    \centering
    \includegraphics[width=\linewidth]{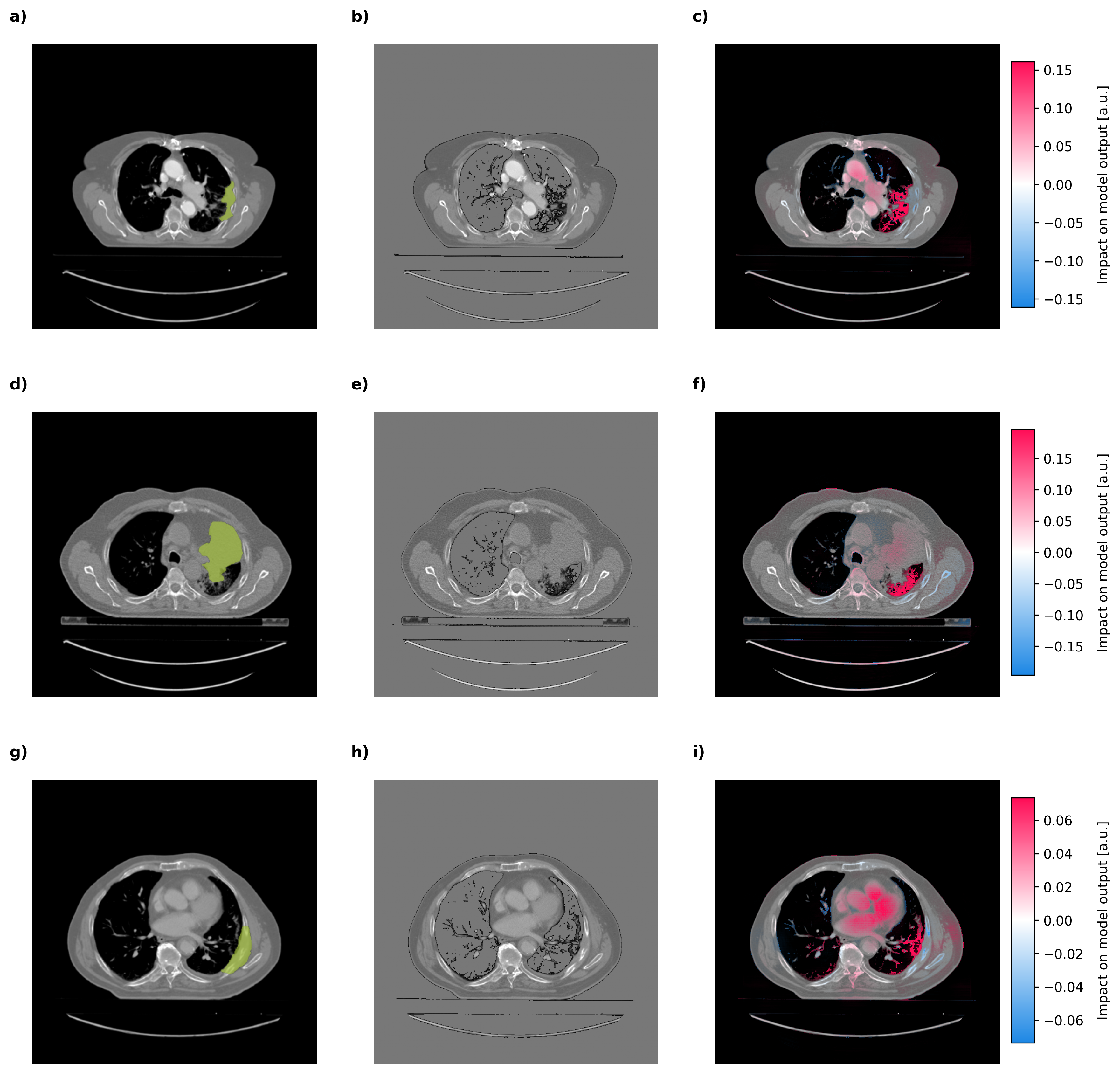}
    \caption{
        \textbf{Co-localization of human-annotated tumor regions with \ac{cnn} attention patterns in high-risk \ac{nsclc} patients.}
        CT scan slices from three representative \ac{nsclc} patients (rows a-c, d-f, and g-i) showing: manual tumor annotations by clinical experts highlighted in green (left column: a, d, g); reference CT images with zero-signal preserving contrast enhancement (middle column: b, e, h); and corresponding \ac{shap} value heatmaps (right column: c, f, i) from the \ac{cnn} trained to cluster images based on patient survival time.
        The \ac{cnn} was trained without access to human tumor annotations.
        \ac{shap} values represent pixel-level importance for the model's survival-based predictions.
        Color scale bars indicate \ac{shap} value magnitude (arbitrary units), representing increased model attention in colored areas.
        Note the remarkable co-localization between expert-identified tumor regions and areas of peak \ac{cnn} attention, demonstrating that the survival prediction model independently learned to focus on clinically relevant tumor features.
        All shown data were derived from the testing partitions of the 5 cross-validation-folds and are entirely based on unseen patients.
    }
    \label{fig:lung_shap_b2b_high_risk}
\end{figure*}

\begin{figure*}[th!]
    \centering
    \includegraphics[width=\linewidth]{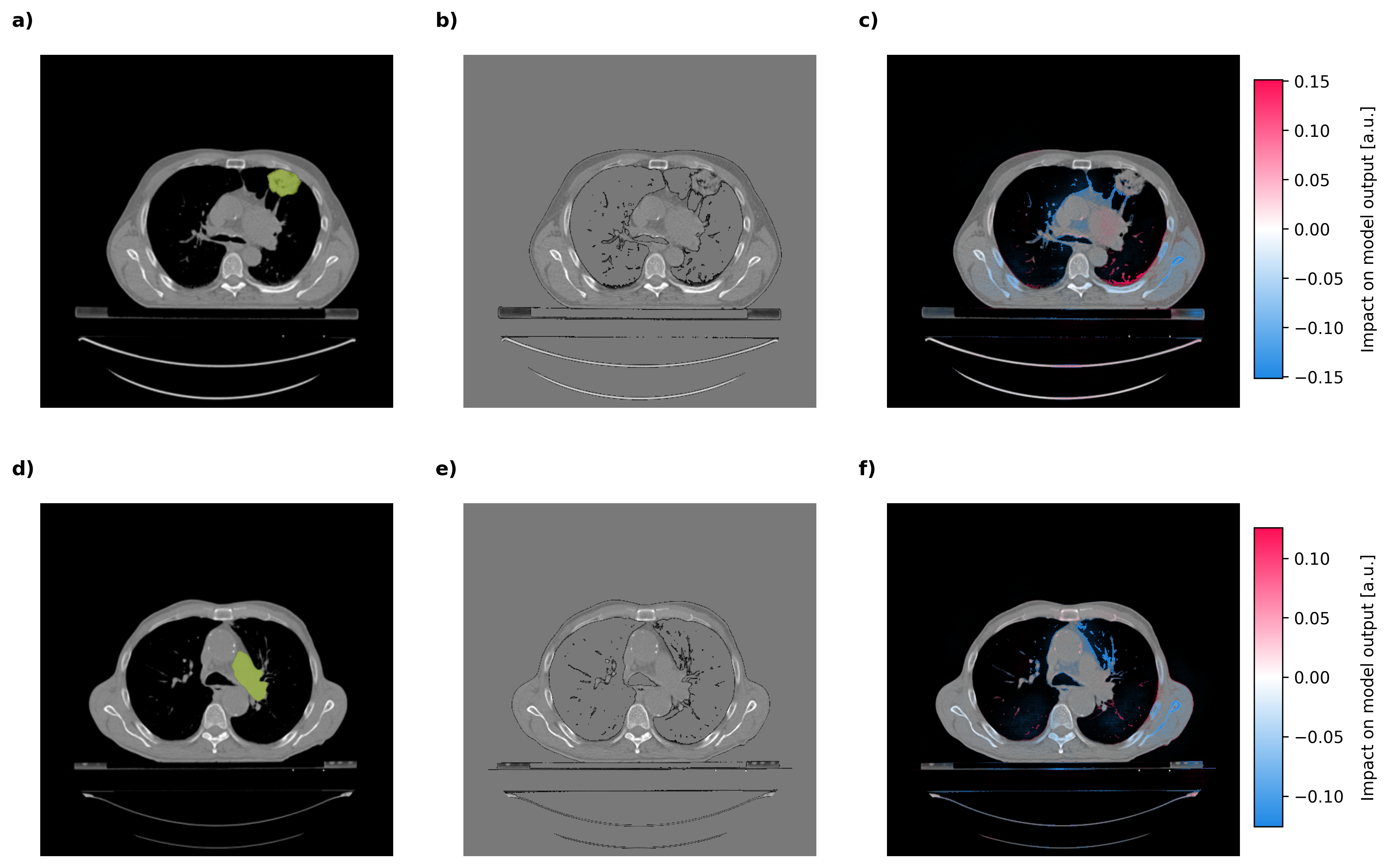}
    \caption{
        \textbf{Co-localization of human-annotated tumor regions with \ac{cnn} attention patterns in low-risk \ac{nsclc} patients.}
        CT scan slices from two representative low-risk \ac{nsclc} patients (rows a-c and d-f) showing: manual tumor annotations by clinical experts highlighted in green (left column: a, d); reference CT images with zero-signal preserving contrast enhancement (middle column: b, e); and corresponding \ac{shap} value heatmaps (right column: c, f) from the \ac{cnn} trained to cluster images based on patient survival time.
        The \ac{cnn} was trained without access to human tumor annotations.
        \ac{shap} values represent pixel-level importance for the model's survival-based predictions.
        Color scale bars indicate \ac{shap} value magnitude (arbitrary units), representing increased model attention in colored areas.
        In contrast to high-risk patients, these low-risk cases demonstrate different attention patterns from the \ac{cnn}, with predominantly negative (blue) \ac{shap} values in tumor regions.
        This suggests the model identifies distinct radiographic features in tumors associated with better prognosis, highlighting the \ac{cnn}'s ability to differentiate survival-relevant characteristics beyond simple tumor detection.
        All shown data were derived from the testing partitions of the 5 cross-validation-folds and are entirely based on unseen patients.
    }
    \label{fig:lung_shap_b2b}
\end{figure*}

Building on the successful application of our method to tabular biomarker data in \ac{mm}, we next investigated the versatility of our optimization framework by applying it to complex medical imaging data.
We developed a complementary approach for stratifying \ac{nsclc} patients based solely on \ac{ct} imaging data, maintaining our core methodology while changing the network architecture to a \ac{cnn} to suit this different data modality.

We applied our method to the Lung1 dataset\supercite{aerts_data_2019}, comprising pretreatment \ac{ct} scans from 422 \ac{nsclc} patients with documented clinical outcomes.
This dataset is particularly suitable for validating our approach as it contains a diverse range of tumor phenotypes and has been extensively studied for prognostic radiomics signatures\supercite{aerts_decoding_2014}, providing a robust benchmark for our methodology.

We trained a custom \ac{cnn} on 2D axial slices of the thoracic \ac{ct} scans to cluster patients into two risk groups (Figure \ref{fig:lung_cluster_fig}a).
Critically, our approach required no manual tumor segmentation or sophisticated preprocessing beyond basic image normalization (see methods section \textit{\nameref{sec:patient_characteristics}}).
The \ac{cnn} was trained using the same \textit{PartialMultivariateLogrankLoss} function employed in our previous experiments, optimizing for maximum between-group survival heterogeneity without any prior knowledge of malignancy, tumor identity, or location.
This \("\)raw-data\("\) approach represents a significant deviation from traditional radiomics methods that typically require extensive manual preprocessing and feature engineering.

Like before, we partitioned the dataset using 5-fold cross-validation and would exclusively evaluate model performance in unseen patients belonging to the withheld testing partitions (Figure \ref{fig:blood_cluster_fig}b-d).
The resulting patient stratification achieved a highly significant separation of survival curves ($p=2.4\times10^{-7}$, Figure \ref{fig:lung_cluster_fig}b), with Cluster 0 showing markedly poorer outcomes compared to Cluster 1.
The median survival times for cluster 0 and cluster one were 446 days and 806 days, respectively, demonstrating robust prognostic performance comparable to the performance reported in the original radiomics analysis by Aerts et al.\supercite{aerts_decoding_2014}, where median survival times were approximately 400 days and 750 days respectively, despite our method requiring substantially less manual preprocessing and feature engineering.

To understand how the model made its decisions, we performed post-hoc \ac{shap} explainability analyses (Figure \ref{fig:lung_cluster_fig}c-d).
Remarkably, despite being trained without any tumor annotations or location information, the model's attention was predominantly focused on the intrapulmonary region of the thorax, specifically on areas containing tumor infiltrates.
This finding demonstrates our methods ability to train models to autonomously identify clinically relevant patient features even in data modalities as complex as \ac{ct} imaging data.

We further validated this observation by comparing the \ac{cnn}'s attention patterns with human expert annotations of tumor masses (Figure \ref{fig:lung_shap_b2b_high_risk}).
These annotations are available in the Lung1 dataset but were deliberately withheld from the model during training.
The comparison revealed robust co-localization between expert-identified tumor regions and areas of peak \ac{cnn} attention, particularly growth patterns in the immediate tumor environments.
This notable alignment confirms that our unsupervised approach successfully learned to identify clinically relevant tumor features without explicit guidance.

Intriguingly, detailed analysis of the attention patterns revealed that the model does not simply recognize the tumor mass directly, but rather focuses on the branching infiltrative tissue originating from the tumors (Figure \ref{fig:lung_shap_b2b_high_risk}c,f,i).
This aligns with clinical understanding of \ac{nsclc} progression, where infiltrative growth patterns and adjacent \ac{ila}\supercite{washko_identification_2010} are associated with more aggressive disease\supercite{ma_relationship_2017, iwasawa_computer-aided_2019, axelsson_associations_2020}.
In high-risk patients, these infiltrative regions display predominantly positive (red) \ac{shap} values, indicating their association with poorer outcomes.

Conversely, in patients classified as low-risk (Figure \ref{fig:lung_shap_b2b}), the model maintains its focus on tumor-adjacent tissue with the lesser branching in tumor morphology resulting in predominantly negative (blue) \ac{shap} values.
This suggests the model identifies specific radiographic features associated with better prognosis, such as more contained growth patterns and less extensive infiltration.
The contrast between attention patterns in high-risk versus low-risk patients demonstrates the model's sophisticated ability to differentiate survival-relevant characteristics beyond simple tumor detection.

These findings are particularly remarkable considering the entirely unsupervised nature of our approach.
Without any prior knowledge of tumor biology or radiographic features, the model autonomously learned to recognize higher degrees of branching infiltrates as risk factor in \ac{nsclc} patients, which is a known adverse prognostic marker for aggressively spreading carcinoma\supercite{ma_relationship_2017, iwasawa_computer-aided_2019}.

        \section{Discussion}\label{sec:discussion}
        In this study, we introduced a novel optimization framework for training neural networks to identify prognostically distinct patient subgroups across diverse cancer types and data modalities.
Our approach addresses a fundamental challenge in oncology: how to leverage the wealth of available patient data to uncover clinically meaningful risk factors without relying on extensive manual feature engineering.
By formalizing the core optimization problem of survival-based clustering without prior knowledge of class labels, we have proposed a universal framework that enables (1) optimal stratification of patients based into prognostically distinct subgroups (2) flexibly across diverse clinical input modalities and (3) irrespective of model architecture.

In this context, it is important to note, that primary scope of this research was to develop novel methodology for training neural networks to be employed in cancer care and for uncovering novel risk factors, rather than to create a specific clinical application.
This distinction is relevant, as our goal was to establish a generalizable framework that could serve as a foundation for future application-specific developments across oncology.

Our systematic validation across four distinct scenarios | application to synthetic data (both tabular- and image-based), as well as real-world biomarker data of \ac{mm} patients and \ac{ct} imaging data of \ac{nsclc} patients | demonstrates the versatility and effectiveness of our approach.
Moreover, in the real-world applications, our method successfully identified prognostically relevant patterns with minimal to no preprocessing and without prior knowledge of established risk factors.

\subsection{Addressing the challenge of unsupervised risk stratification}
A central challenge in developing unsupervised approaches of any kind but especially for patient stratification is the absence of definitive ground truth to validate obtained clustering against\supercite{ofer_interfeat_2025}.
In clinical applications, we can assess statistical separation of resulting survival distributions or if these are based on clinically interpretable features, as we did in our applications to \ac{mm} and \ac{nsclc} data, but cannot definitively determine whether the identified groups represent the optimal stratification.
Our controlled simulation experiments with synthetic data provided a crucial validation step for our framework by creating a ground truth association between specific input features and survival outcomes, allowing us to systematically evaluate the frameworks ability to enable models to recover these predefined relationships.
The strong performance of our approach in these controlled settings (achieving AUROC and AUPRC values above 0.90 for both tabular and image data) provided the necessary evidence for our optimization framework being able to readily identify features that are associated with prognostically distinct patient subgroups, offering confidence in its application in real-world clinical settings.

\subsection{Clinical relevance in \ac{mm}}
Our application to \ac{mm} biomarker data revealed particularly promising results with immediate clinical relevance.
Our model's performance in \ac{mm} risk stratification is encouraging, as it recapitulates known clinical associations while achieving similar performance to established staging systems using only routine blood parameters.
The model's ability to approach the performance of the \ac{riss} and IrMMa system (c-index 0.644 vs. 0.61\supercite{cho_evaluation_2024, brieghel_real-world_2025} vs. 0.687\supercite{maura_genomic_2024} respectively) is notable, given that the \ac{riss} incorporates cytogenetic abnormalities\supercite{palumbo_revised_2015}, while IrMMa relies on genomic data\supercite{maura_genomic_2024}, which are among the strongest known prognostic factors in \ac{mm}.
This finding suggests that sophisticated machine learning approaches can extract comparable or even superior prognostic information from routine blood parameters alone, potentially reducing the need for specialized testing in some clinical scenarios and enabling risk stratification in resource-limited settings where cytogenetic testing may be unavailable or non-economical.

Moreover, while the role of well-known risk factors in established clinical staging systems including \ac{b2m}, \ac{alb}, \ac{cr}, and \ac{ldh}\supercite{greipp_international_2005} were reproduced, the model also identified nuanced relationships between biomarkers that extend beyond conventional clinical interpretations.
Here, the distribution of \ac{sfl}s across the clusters also demonstrates the potential our approach, with the high-risk cluster showing significantly elevated levels compared to the favorable-risk cluster (Figure \ref{fig:blood_cluster_fig}c).
This pattern reflects the known association between elevated free light chains and aggressive disease biology\supercite{li_correlation_2024, benson_early_2020}, including higher rates of renal impairment and faster disease progression\supercite{jin_light_2021, li_correlation_2024}.

\subsection{Utility in medical imaging}
Our application to CT imaging data in \ac{nsclc} patients further validates the utility of our framework.
Without any tumor annotations or location information during training, our method was able to autonomously learn to focus on clinically relevant regions, with attention patterns that showed notable co-localization with expert-identified tumor areas, while focusing on prognostically relevant tumor morphology.

This ability to autonomously recognize infiltrative tumor tissue growths and \ac{ila}, which are well-known risk factors in \ac{nsclc}\supercite{washko_identification_2010, drakopanagiotakis_lung_2024, hida_interstitial_2021, zhu_newly_2023, axelsson_associations_2020}, represents a sophisticated level of feature learning that goes beyond simple tumor detection.
This finding is particularly significant as it demonstrates that survival outcomes alone can guide neural networks to discover clinically meaningful imaging biomarkers without explicit supervision.

To extract such highly interpretable and clinically relevant features in an unsupervised manner represents a significant advancement in the application of artificial intelligence in cancer\supercite{waldstein_unbiased_2020, pai_foundation_2024, shen_deep_2017}.
Here, our approach both achieves robust prognostic performance and provides explainable results that could complement clinical decision-making and potentially reveal novel imaging biomarkers.
These results powerfully demonstrate the utility of our proposed method for uncovering prognostically significant patterns across diverse data modalities, with particularly promising applications in medical imaging where manual feature extraction is time-consuming and potentially subject to human bias\supercite{parmar_robust_2014, shen_deep_2017}.

\subsection{Outlook and potential improvements}
Despite the promising findings, several limitations of our study warrant discussion.

First, it is important to emphasize that, in the research presented here, we were not aiming at training a model for a specific application but to develop and validate a unifying framework for broad applicability of artificial intelligence in cancer care.
While, in the work we present here, we adhered to best practices for model validation by cross-validating results with unseen patients exclusively, we acknowledge, however, that if our approach were employed for the discovery of novel risk factors, it would be essential to validate such risk factors in independent cohorts.
Notably, we could identify well-known clinical parameters reflecting disease burden in \ac{mm} and radiographic features in \ac{nsclc} that are well-known to be associated with poor outcomes, which provides the necessary evidence of our methodology's generalizability and powerfully demonstrates its utility.
Hence, the validation of any particular model on an external dataset was outside the scope of our research endeavors, as we were not aiming at developing a specific model but rather a flexible framework, agnostic of application-specific modeling requirements.

Second, our approach currently relies on post-hoc interpretability methods like SHAP analysis to understand model decisions.
Future work that focuses on a specific modeling task could explore integrating interpretability directly into the architecture of the respective models to avoid the need for post-hoc analyses and provide more readily available explanations of the model outputs.

Third, in our current implementation, the optimal number of risk groups must be specified a priori.
Determining the optimal number of clusters is a key challenge in unsupervised learning.
While we drew on clinical knowledge to inform our choices for cluster numbers in the presented use cases, determining the optimal number of clusters in a setting where clinical knowledge is limited should be guided by through elbow methods\supercite{thorndike_who_1953}, a dirichlet process\supercite{sammut_dirichlet_2011}, or other similar techniques.

Finally, while in the applications we presented here, each data modality was treated separately, integrating multiple data types (e.g., imaging, genomics, and clinical parameters) into a unified multi-modal model could potentially enhance prognostic accuracy and provide more comprehensive patient characterization, which is a promising direction for future work.

\subsection{Broader implications for precision oncology}
Traditional cancer classification systems like \ac{riss} and \ac{tnm} staging, while foundational to clinical practice, struggle to adapt to the complexities of modern oncology\supercite{kwon_development_2025}.
These systems typically rely on a limited number of predefined risk factors, which may not capture the full spectrum of disease heterogeneity or the complex interactions between different prognostic factors.

Our approach represents a paradigm shift in cancer risk stratification by enabling the discovery of prognostically relevant patterns directly from patient data without prior assumptions or feature selection.
This data-driven approach has the potential to identify novel risk factors and patient subgroups that might be missed by conventional staging systems\supercite{ofer_interfeat_2025, waldstein_unbiased_2020, pai_foundation_2024}.

This flexibility is particularly valuable in the era of precision medicine, where treatment decisions increasingly depend on integrating diverse data types to characterize individual patients' disease biology\supercite{kwon_development_2025}.
By enabling the identification of prognostically distinct patient subgroups across different data modalities, our framework could help bridge the gap between the growing wealth of patient data and actionable clinical insights.

\subsection{Conclusion}
In conclusion, our study introduces a novel optimization framework for training neural networks to identify prognostically distinct patient subgroups across diverse cancer types and data modalities.
By directly optimizing for survival heterogeneity, our approach enables the discovery of clinically meaningful patient stratifications without prior knowledge of established risk factors or extensive manual feature engineering.

%The successful application of our methodology across synthetic data, \ac{mm} biomarkers, and \ac{nsclc} imaging demonstrates its versatility and potential clinical utility.
%Our framework represents a step forward in enabling personalized, data-driven treatment decisions, with the potential to evolve alongside ongoing advances in cancer research and patient care.
As we continue to refine and expand this methodology, we envision its integration into clinical decision support systems that can help oncologists navigate the growing complexity of cancer diagnosis, prognosis, and treatment selection.
Thus, by providing a unified approach to patient stratification across different cancer types and data modalities, our framework contributes to the broader goal of precision oncology: matching patients with the most effective treatments at the right time\supercite{de_jong_towards_2021, qiu_deep_2025}.

        \section{Methods}\label{sec:methods}
        \subsection{Ethics Statement}
The CoMMpass study was funded by the \ac{mm} Research Foundation (MMRF) and conducted in line with the Declaration of Helsinki.
Approval for the study was granted to the MMRF by the second panel of the Western Institutional Review Board.
The CoMMpass study data is publicly deposited by the MMRF in anonymized form (refer to section \textit{\nameref{sec:data_availability}}).
Participants were not paid for their involvement in the CoMMpass study and only joined the study after giving their written informed consent.

The Lung1 study was conducted according to Dutch law and in line with the Declaration of Helsinki.
Approval for the Lung1 study was granted to Aerts et al.\supercite{aerts_decoding_2014} by the local ethics committee at Maastricht University Medical Center (MUMC1), Maastricht, The Netherlands.
The Lung1 study data is publicly deposited by Aerts et al. at The Cancer Imaging Archive in anonymized form (refer to section \textit{\nameref{sec:data_availability}}).

All research in our study was carried out retrospectively with the publicly available data and in compliance with the Declaration of Helsinki.

\subsection{Patient characteristics and data preprocessing}\label{sec:patient_characteristics}
For model training and validation on clinical tabular data, we retrieved \ac{mm} laboratory patient data from the CoMMpass study (NCT01454297) database version IA21.
For \ac{mm} disease kinetics modeling, we adopted the parameter selection from our previous work\supercite{ferle_predicting_2025}, which identified key biomarkers based on their clinical relevance: \ac{hb} (anemia prevalence), \ac{ca} (bone metabolism and hypercalcemia), \ac{cr} (renal complications), \ac{wbc} (treatment-related leukopenia and infection risk), MM-specific disease markers (\ac{mpr}, \ac{sfl}), and prognostic indicators (\ac{ldh}, \ac{alb}, \ac{b2m}). Furthermore, we have chosen these parameters because they represent standard laboratory measurements in MM care and are therefore easily accessible in routine clinical practice.
To assess model performance we employed a 5-fold cross validation scheme, where we trained the model on 80\% of patients and evaluated its performance on the remaining 20\% unseen patients across five different partitions of the data.
The only patients excluded were those for which no survival documentation was available, which left 998 patients in total, resulting to 799 patients for model training and 199 patients for validation in each of the five cross validation folds respectively.
Missing values for lab values where imputed using the mean value of each respective parameter.
Importantly, imputation parameters were derived from the training data exclusively to avoid bias in the evaluation of model performance.
All laboratory parameters were subject to a feature-wise standard normalization (mean = 0, standard deviation = 1).
Likewise, normalization parameters were derived from the training data exclusively to avoid bias.
Table \ref{tab:mm_patient_characteristics} conveys an overview of the relevant patient characteristics.
\begin{table*}[t]
    \small
    \centering
    \caption{Patient characteristics table for \ac{mm} cohort separated by cross-validation ($k^{th}$) fold.}
    \label{tab:mm_patient_characteristics}
    \resizebox{\textwidth}{!}{%
        \begin{tabular}{lrrrrr}
            \toprule
            & \textit{k}=1 (\textit{N}=199) & \textit{k}=2 (\textit{N}=199) & \textit{k}=3 (\textit{N}=199)               & \textit{k}=4 (\textit{N}=199)               & \textit{k}=5 (\textit{N}=199)                \\
            \midrule
            Age (years)                                  & 64.8 $\pm$ 11.6               & 63.6 $\pm$ 10.5               & 64.5 $\pm$ 10.5               & 63.0 $\pm$ 9.7                & 63.4 $\pm$ 10.8               \\
            \midrule
            \multicolumn{6}{l}{\textbf{ISS Stage}} \\
            ISS 1                                        & 65 (32.7\%)                   & 75 (37.7\%)                   & 67 (33.7\%)                   & 63 (31.7\%)                   & 62 (31.2\%)                   \\
            ISS 2                                        & 71 (35.7\%)                   & 59 (29.6\%)                   & 60 (30.2\%)                   & 69 (34.7\%)                   & 87 (43.7\%)                   \\
            ISS 3                                        & 56 (28.1\%)                   & 56 (28.1\%)                   & 64 (32.2\%)                   & 62 (31.2\%)                   & 49 (24.6\%)                   \\
            \midrule
            \multicolumn{6}{l}{\textbf{Stem Cell Transplant}} \\
            No SCT                                       & 98 (49.2\%)                   & 101 (50.8\%)                  & 96 (48.2\%)                   & 77 (38.7\%)                   & 98 (49.2\%)                   \\
            SCT                                          & 101 (50.8\%)                  & 98 (49.2\%)                   & 103 (51.8\%)                  & 122 (61.3\%)                  & 101 (50.8\%)                  \\
            \midrule
            \multicolumn{6}{l}{\textbf{Treatment Class}} \\
            Bortezomib-based                             & 49 (24.6\%)                   & 39 (19.6\%)                   & 52 (26.1\%)                   & 31 (15.6\%)                   & 43 (21.6\%)                   \\
            Carfilzomib-based                            & 1 (0.5\%)                     & 2 (1.0\%)                     & 6 (3.0\%)                     & 6 (3.0\%)                     & 0 (0.0\%)                     \\
            IMIDs-based                                  & 11 (5.5\%)                    & 15 (7.5\%)                    & 8 (4.0\%)                     & 16 (8.0\%)                    & 16 (8.0\%)                    \\
            Combined IMIDs/carfilzomib-based             & 18 (9.0\%)                    & 14 (7.0\%)                    & 11 (5.5\%)                    & 19 (9.5\%)                    & 13 (6.5\%)                    \\
            Combined bortezomib/IMIDs-based              & 115 (57.8\%)                  & 121 (60.8\%)                  & 114 (57.3\%)                  & 114 (57.3\%) & 115 (57.8\%) \\
            Combined bortezomib/IMIDs/carfilzomib-based  & 5 (2.5\%)                     & 7 (3.5\%)                     & 8 (4.0\%)                     & 12 (6.0\%) & 11 (5.5\%) \\
            Combined bortezomib/carfilzomib-based        & 0 (0.0\%)                     & 0 (0.0\%)                     & 0 (0.0\%)                     & 1 (0.5\%)                     & 1 (0.5\%)                     \\
            Combined daratumumab/IMIDs/carfilzomib-based & 0 (0.0\%)                     & 1 (0.5\%)                     & 0 (0.0\%)                     & 0 (0.0\%) & 0 (0.0\%) \\
            \midrule
            \multicolumn{6}{l}{\textbf{Laboratory Values}} \\
            Hemoglobin [mmol/L]                          & 6.64 $\pm$ 1.21               & 6.79 $\pm$ 1.13               & 6.64 $\pm$ 1.22               & 6.61 $\pm$ 1.17       & 6.63 $\pm$ 1.24        \\
            Calcium [mmol/L]                             & 2.35 $\pm$ 0.25               & 2.38 $\pm$ 0.31               & 2.42 $\pm$ 0.29               & 2.39 $\pm$ 0.33       & 2.35 $\pm$ 0.30        \\
            Creatinine [$\mu$mol/L]                      & 119.61 $\pm$ 111.17           & 129.38 $\pm$ 120.74           & 125.50 $\pm$ 94.44    & 131.93 $\pm$ 172.66 & 114.64 $\pm$ 98.38 \\
            LDH [$\mu$kat/L]                             & 3.40 $\pm$ 1.79               & 3.68 $\pm$ 2.43               & 3.47 $\pm$ 1.96               & 3.51 $\pm$ 2.92       & 3.29 $\pm$ 1.95        \\
            Albumin [g/L]                                & 35.32 $\pm$ 6.42              & 36.10 $\pm$ 6.45              & 37.12 $\pm$ 6.22              & 34.94 $\pm$ 6.31      & 35.62 $\pm$ 6.20       \\
            $\beta$-2-Microglobulin [mg/L]               & 5.20 $\pm$ 4.80               & 4.96 $\pm$ 4.47               & 5.45 $\pm$ 4.99               & 5.39 $\pm$ 5.10       & 4.88 $\pm$ 4.03        \\
            M-protein [g/dL]                             & 2.63 $\pm$ 2.17               & 2.68 $\pm$ 2.38               & 2.77 $\pm$ 2.39               & 2.98 $\pm$ 2.30       & 2.80 $\pm$ 2.37        \\
            Lambda light chain [mg/dL]                   & 130.04 $\pm$ 830.95           & 119.23 $\pm$ 516.70           & 96.50 $\pm$ 322.00    & 126.32 $\pm$ 596.15 & 87.39 $\pm$ 268.72 \\
            Kappa light chain [mg/dL]                    & 315.85 $\pm$ 1810.64          & 127.19 $\pm$ 291.28           & 244.61 $\pm$ 819.26   & 216.74 $\pm$ 710.28 & 2743.83 $\pm$ 32459.86 \\
            White blood cell count [10$^9$/L]            & 6.11 $\pm$ 2.35               & 6.53 $\pm$ 2.93               & 6.25 $\pm$ 2.20       & 6.24 $\pm$ 2.42 & 6.50 $\pm$ 3.33 \\
            \midrule
            \multicolumn{6}{l}{\textbf{Survival Outcomes}} \\
            Median survival [days (IQR)]                      & 1749 (632|2652)               & 1770 (610|2513)               & 1986 (780|2580)               & 1944 (796|2559) & 1690 (789|2512) \\
            Censoring rate                               & 54.3\%                        & 60.8\%                        & 58.3\%                        & 61.8\%                        & 53.3\%                        \\
            \bottomrule
        \end{tabular}%
    }
\end{table*}

For model training and validation on clinical imaging data, we retrieved \ac{ct} scans of \ac{nsclc} patients from Lung1 study dataset version 4 (2020)\supercite{aerts_data_2019}.
The Lung1 study dataset includes 422 patients with documented clinical outcomes and is publicly available at the \textit{Cancer Imaging Archive}.
The inclusion criteria for the study were a confirmed primary tumour in the \ac{ct} scan and curative-intent treatment\supercite{aerts_decoding_2014}.
The unaltered \ac{ct} scans were normalized for model training using \ac{zpce}.
This normalization subjects tissue pixel intensities to a standard normalization (mean = 0, standard deviation = 1), while background pixels remain unchanged, which effectively makes subtle structural differences more detectable by amplifying intensity variations within tissue regions.
Like in the application to \ac{mm}, model performance was assessed using a 5-fold cross validation scheme, where we trained the model on 80\% of patients and evaluated its performance on the remaining 20\% unseen patients across five different partitions of the data.
We did not exclude any patients from the dataset in our analysis, which resulted in 336 patients for training and 84 patients for validation in each of the five cross validation folds.
Furthermore, the training data was augmented by decomposing 3D \ac{ct} scans into 2D axial slices, selecting 2 to 3 slices per patient that contained visible \ac{nsclc} tissue, yielding a total of 1,118 slices for model training.
Validation was done on a single slice per patient which contained the largest cross-section of \ac{nsclc} tissue.
Table \ref{tab:nsclc_patient_characteristics} conveys an overview of the relevant patient characteristics.
\begin{table*}[t]
    \small
    \centering
    \caption{Patient characteristics table for NSCLC cohort separated by cross-validation ($k^{th}$) fold.}
    \label{tab:nsclc_patient_characteristics}
    \resizebox{\textwidth}{!}{%
        \begin{tabular}{lrrrrr}
            \toprule
            & \textit{k}=1 (\textit{N}=84) & \textit{k}=2 (\textit{N}=84) & \textit{k}=3 (\textit{N}=84) & \textit{k}=4 (\textit{N}=84) & \textit{k}=5 (\textit{N}=84) \\
            \midrule
            Age (years) & 68.3 $\pm$ 10.0 & 69.2 $\pm$ 9.6 & 68.3 $\pm$ 10.1 & 67.8 $\pm$ 10.1 & 66.6 $\pm$ 10.6 \\
            \midrule
            \multicolumn{6}{l}{\textbf{T Stage}} \\
            T1 & 16 (19.0\%) & 17 (20.2\%) & 25 (29.8\%) & 21 (25.0\%) & 14 (16.9\%) \\
            T2 & 31 (36.9\%) & 29 (34.5\%) & 30 (35.7\%) & 31 (36.9\%) & 35 (42.2\%) \\
            T3 & 13 (15.5\%) & 12 (14.3\%) & 8 (9.5\%) & 9 (10.7\%) & 10 (12.0\%) \\
            T4 & 24 (28.6\%) & 24 (28.6\%) & 21 (25.0\%) & 23 (27.4\%) & 24 (28.9\%) \\
%            T5 & 0 (0.0\%) & 2 (2.4\%) & 0 (0.0\%) & 0 (0.0\%) & 0 (0.0\%) \\
            \midrule
            \multicolumn{6}{l}{\textbf{N Stage}} \\
            N0 & 38 (45.2\%) & 27 (32.1\%) & 38 (45.2\%) & 37 (44.0\%) & 29 (34.5\%) \\
            N1 & 5 (6.0\%) & 5 (6.0\%) & 7 (8.3\%) & 3 (3.6\%) & 2 (2.4\%) \\
            N2 & 28 (33.3\%) & 31 (36.9\%) & 22 (26.2\%) & 31 (36.9\%) & 29 (34.5\%) \\
            N3 & 13 (15.5\%) & 20 (23.8\%) & 16 (19.0\%) & 13 (15.5\%) & 23 (27.4\%) \\
            \midrule
            \multicolumn{6}{l}{\textbf{M Stage}} \\
            M0 & 84 (100.0\%) & 82 (97.6\%) & 83 (98.8\%) & 84 (100.0\%) & 82 (97.6\%) \\
            M1 & 0 (0.0\%) & 1 (1.2\%) & 0 (0.0\%) & 0 (0.0\%) & 0 (0.0\%) \\
%            M2 & 0 (0.0\%) & 0 (0.0\%) & 0 (0.0\%) & 0 (0.0\%) & 0 (0.0\%) \\
%            M3 & 0 (0.0\%) & 1 (1.2\%) & 1 (1.2\%) & 0 (0.0\%) & 2 (2.4\%) \\
            \midrule
            \multicolumn{6}{l}{\textbf{Overall Stage}} \\
            I & 18 (21.4\%) & 13 (15.7\%) & 24 (28.6\%) & 23 (27.4\%) & 15 (17.9\%) \\
            II & 10 (11.9\%) & 11 (13.3\%) & 9 (10.7\%) & 3 (3.6\%) & 7 (8.3\%) \\
            IIIa & 20 (23.8\%) & 24 (28.9\%) & 17 (20.2\%) & 29 (34.5\%) & 21 (25.0\%) \\
            IIIb & 36 (42.9\%) & 35 (42.2\%) & 34 (40.5\%) & 29 (34.5\%) & 41 (48.8\%) \\
            \midrule
            \multicolumn{6}{l}{\textbf{Histology}} \\
            Adenocarcinoma & 11 (14.7\%) & 11 (14.5\%) & 9 (12.3\%) & 7 (9.5\%) & 13 (16.2\%) \\
            Large cell & 20 (26.7\%) & 22 (28.9\%) & 30 (41.1\%) & 22 (29.7\%) & 20 (25.0\%) \\
            NOS & 15 (20.0\%) & 14 (18.4\%) & 5 (6.8\%) & 10 (13.5\%) & 17 (21.2\%) \\
            Squamous cell carcinoma & 29 (38.7\%) & 29 (38.2\%) & 29 (39.7\%) & 35 (47.3\%) & 30 (37.5\%) \\
            \midrule
            \multicolumn{6}{l}{\textbf{Survival Outcomes}} \\
            Median survival [days (IQR)] & 516 (248|1477) & 563 (286|1411) & 563 (260|1526) & 528 (280|1034) & 675 (277|1510) \\
            Censoring rate & 10.7\% & 11.9\% & 13.1\% & 8.3\% & 13.1\% \\
            \bottomrule
        \end{tabular}%
    }
\end{table*}
\subsection{Development of a differentiable optimization criterion}\label{sec:eqns}
Consider a survival analysis setting with \( k \) distinct groups comprising \( n \) individuals.
Initially, let \( g_i \in \{1, 2, 3, ..., k\} \) represent the definitive group membership of the \( i \)-th individual.
The Multivariate Logrank statistic \( L \), which quantifies the heterogeneity of survival curves across groups, can be constructed as follows:
For each group \( g \) and unique event time \( t_j \), the observed events \( O_{g,j} \) are calculated using:
\begin{equation}
    O_{g,j} = \sum_{i \in D(t_j)} \delta_{g_i, g} \text{\hspace{10pt} with \hspace{10pt}}
    \delta_{g_i, g} = \begin{cases}
                          1 & \text{if } g_i = g \\
                          0 & \text{if } g_i \neq g
    \end{cases}
    \label{eq:observed_events}
\end{equation}
where \( D(t_j) \) represents the set of individuals experiencing the event at time \( t_j \), and \( \delta_{g_i, g} \) indicates group membership.
The corresponding expected events \( E_{g,j} \) for group \( g \) at time \( t_j \) are given by:
\begin{equation}
    E_{g,j} = \frac{R_g(t_j)}{|R(t_j)|} \cdot d_j
    \label{eq:expected_events}
\end{equation}
Here, \( R_g(t_j) \) denotes the number of at-risk individuals in group \( g \) at time \( t_j \) , \( |R(t_j)| \) represents the total at-risk population across all groups at time \( t_j \), and \( d_j \) accounts for the total number of events at time \( t_j \), incorporating censoring information.
The multivariate log-rank statistic is then:
\begin{equation}
    L = \mathbf{Z}^T \mathbf{V}^{-1} \mathbf{Z}
    \label{eq:logrank_statistic}
\end{equation}
where \( \mathbf{Z} = [Z_1, Z_2, ..., Z_{k}]^T \) with \( Z_g = \sum_j (O_{g,j} - E_{g,j}) \) and \(  \mathbf{V} \) represents the full variance-covariance matrix that accounts for correlations between groups, calculated using the standard hypergeometric variance formula for the log-rank test\supercite{mantel_evaluation_1966, peto_asymptotically_1972}.

To leverage modern machine learning based clustering techniques, we can generalize this framework to accommodate soft class assignments.
This is achieved by redefining \( g_i \) as a categorical probability distribution over the class labels:
\begin{equation}
    g_i = [p_{i1}, p_{i2}, \dots, p_{ik}]
    \text{\hspace{10pt} with \hspace{10pt}}
    \int g_i = 1
    \label{eq:soft_class_assignments}
\end{equation}
where  \( p_{ig} \) reflects the probability of individual \( i \) to belong to group \( g \).
Here, \( g_i \) is modeled as a function \( g_i(\vec{x}_i, \theta) \) of the features \( \vec{x}_i \) of individual \( i \) and model parameters \( \theta \) of a neural network.
This probabilistic framework generalizes the observed events calculation to partial events:
\begin{equation}
    O_{g,j} = \sum_{i \in D(t_j)} p_{ig}
    \label{eq:soft_observed_events}
\end{equation}
where \( D(t_j) \) remains the total set of individuals experiencing the event at time \( t_j \).
Similarly, the risk set for group \( g \) at time \( t_j \) becomes a weighted sum of probabilities:
\begin{equation}
    R_g(t_j) = \sum_{i \in R(t_j)} p_{ig}
    \label{eq:soft_risk_set}
\end{equation}
where \( R(t_j) \) is then a set of partial individuals at risk at time \( t_j \) and allows for calculation of expected events \( E_{g,j} \) for group \( g \) at time \( t_j \) identically to eqn. \ref{eq:expected_events}.

The variance-covariance matrix \( \mathbf{V} \) and the observed minus expected events \( \mathbf{Z} \) also extend naturally to the probabilistic setting, so that the log-rank loss \( L \) can be computed identically to eqn. \ref{eq:logrank_statistic} based on the evaluation of eqns. \ref{eq:soft_observed_events}, \ref{eq:soft_risk_set}, and \ref{eq:expected_events}.
This formulation preserves the statistical properties of the log-rank statistic while allowing for soft class assignments.
The key advantage of this probabilistic formulation is that it creates a fully differentiable objective function, which enables gradient-based optimization of the model parameters \( \theta \) through differentiation of \( L \) with respect to \( \theta \) using the chain rule:
\begin{small}
    \begin{equation}
        \frac{\partial L}{\partial \theta} = \sum_{g} \left( \frac{\partial L}{\partial O_{g}} \frac{\partial O_{g}}{\partial g} \frac{\partial g}{\partial \theta} + \frac{\partial L}{\partial E_{g}} \frac{\partial E_{g}}{\partial g} \frac{\partial g}{\partial \theta} + \frac{\partial L}{\partial V} \frac{\partial V}{\partial g} \frac{\partial g}{\partial \theta} \right)
    \end{equation}
\end{small}
so that the model parameters \( \theta \) can then be optimized iteratively based on the gradient of \( L \) by applying established learning rules, i.e.:
\begin{equation}
    \theta \leftarrow \theta + \alpha \frac{\partial L}{\partial \theta} + R(\theta)
    \label{eq:theta_update}
\end{equation}
As a result, changes in \( \theta \) affect the predicted probabilities \( p_{ig} \), which in turn propagate through the observed events, expected events, and variance calculations to the final log-rank loss.
By virtue of this approach, our method enables training models to map patient features to cluster assignments, allowing direct optimization of distinct patient groups with maximally different survival outcomes.
\subsection{Simulation of synthetic survival data for framework validation}\label{sec:surv_sim}

For survival time simulation, we established three distinct risk groups with different prognostic profiles.

Survival times were generated from Weibull distributions with group-specific parameters chosen to create clinically realistic survival patterns.
The Weibull distribution was selected due to its flexibility in modeling various hazard functions commonly observed in medical data\supercite{plana_cancer_2022, majer_estimating_2022}.
Event times were generated using inverse-transform sampling from the Weibull distribution:
\begin{equation}
    T = \lambda(-\ln(1-U))^{1/\rho},
\end{equation}
where $U$ follows a uniform distribution on $[0, 1]$.
Group-specific parameters were:
\begin{align*}
    \text{Group 0: }&\rho_0=0.539, \lambda_0=3068.812,\\
    \text{Group 1: }&\rho_1=0.898, \lambda_1=5114.687,\\
    \text{and Group 2: }&\rho_2=1.257, \lambda_2=7160.562,
\end{align*}
where $\rho$ represents the shape parameter and $\lambda$ the scale parameter.
The parameters were obtained by using the CoMMpass dataset as a reference, stratifying patients by the ISS stage\supercite{greipp_international_2005}, and fitting Weibull distributions to each group.
Through this process, we obtained a set of three group-specific survival distributions that were based on real-world evidence, which accurately reflected a clinical prognostic setting.
To simulate realistic clinical scenarios with censoring, we introduced random censoring using exponential distribution with scale parameter $10,000$ to mirror censoring rates typical of clinical studies.
Administrative censoring was applied at $4,000$ time units to simulate finite study follow-up periods.

To evaluate and validate our framework's ability to train models to associate optical features with survival outcomes, we utilized a version of the MNIST Handwritten Digits dataset\supercite{e_alpaydin_optical_1998, li_deng_mnist_2012} from the UCI Machine Learning Repository (\url{https://archive.ics.uci.edu}), which contains 8$\times$8 pixel grayscale images of handwritten digits 0-9.
To create a survival analysis setting, we excluded digit 0 and retained digits 1-9.
Each image was normalized using standard scaling to ensure consistent feature magnitudes across the dataset.

Based on digit labels we defined a mapping to one of the three risk groups in a randomized fashion using a fixed permutation.
The final dataset comprised of approximately $5,000$ samples with known group memberships and distinct survival and censoring distributions for each group.
All random number generation was performed with fixed seeds to ensure reproducibility.

\subsection{Model implementation and training}\label{sec:model_training}
All employed models were implemented using the PyTorch\supercite{paszke_pytorch_2019} framework (version 2.5.1) and trained using PyTorchLightning\supercite{falcon_pytorch_2023} (version 2.5.0.post0).

The \ac{mlp} for clustering the synthetic feature vectors associated with survival data was configured with one hidden layer and an output layer:
\begin{align*}
    & \textit{Linear } 3 \mapsto 16\\
    & \textit{Linear } 16 \mapsto 3 + \textit{Softmax}.
\end{align*}
The model was trained at a learning rate of 0.01, for 50 epochs with a mini-batch size of 32, using a uniform weight-decay of 0.01 applied to weights and biases.

The \ac{mlp} for application on real-world clinical biomarker data of \ac{mm} patients was configured with three hidden layers and an output layer:
\begin{align*}
    & \textit{Linear } 10 \mapsto 32\\
    & \textit{Linear } 32 \mapsto 32\\
    & \textit{Linear } 32 \mapsto 32\\
    & \textit{Linear } 32 \mapsto 3 + \textit{Softmax}.
\end{align*}
The model was trained at a learning rate of 0.001, for 20 epochs with a mini-batch size of 32, using a uniform weight-decay of 1 applied to weights and biases.

The \ac{cnn} for application on handwritten digit images associated with synthetic survival data was configured with three convolution-and-pooling blocks followed by a single linear layer with a Softmax activation:
\begin{align*}
    & \textit{Convolution } 1 \mapsto 64 \text{ } (5 \times 5) + \textit{MaxPooling } (2 \times 2)\\
    & \textit{Convolution } 64 \mapsto 32 \text{ } (3 \times 3) + \textit{MaxPooling } (2 \times 2)\\
    & \textit{Convolution } 32 \mapsto 8 \text{ } (5 \times 5) + \textit{MaxPooling } (2 \times 2)\\
    & \textit{Linear } 8 \mapsto 3 + \textit{Softmax}.
\end{align*}
All convolutional layers use zero-padding.
Through the successive convolution and pooling operations, the spatial dimensions (height and width) of the feature maps are progressively reduced until the final 8-channel activation map comprises a one-dimensional vector that can be directly passed to the fully-connected layer.
The model was trained at a learning rate of 0.001, for 20 epochs with a mini-batch size of 64, using a uniform weight-decay of 0.01 applied to weights and biases.

The \ac{cnn} for application on real-world CT imaging data of \ac{nsclc} patients was configured with six convolution-and-pooling blocks followed by a single linear layer with a Softmax activation:
\begin{align*}
    & \textit{Convolution } 1 \mapsto 32 \;(16 \times 16) + \textit{MaxPooling } (4 \times 4)\\
    & \textit{Convolution } 32 \mapsto 8 \;(8 \times 8) + \textit{MaxPooling } (4 \times 4)\\
    & \textit{Convolution } 8 \mapsto 8 \;(8 \times 8) + \textit{MaxPooling } (4 \times 4)\\
    & \textit{Convolution } 8 \mapsto 8 \;(8 \times 8) + \textit{MaxPooling } (2 \times 2)\\
    & \textit{Convolution } 8 \mapsto 4 \;(3 \times 3) + \textit{MaxPooling } (2 \times 2)\\
    & \textit{Convolution } 4 \mapsto 4 \;(2 \times 2) + \textit{MaxPooling } (2 \times 2)\\
    & \textit{Linear } 4 \mapsto 2 + \textit{Softmax}.
\end{align*}
All convolutional layers use zero-padding.
Through the successive convolution and pooling operations, the spatial dimensions (height and width) of the feature maps are progressively reduced until the final 4-channel activation map comprises a one-dimensional vector that can be directly passed to the fully-connected layer.
The model was trained at a learning rate of 0.001 for 100 epochs with mini-batches of 32 images, using a uniform weight-decay of 0.1 applied to weights and biases.

Model hyperparameters and training parameters were determined through grid search.
All models were trained on our custom \textit{PartialMultivariateLogRankLoss} using a class-imbalance penalty of 0.1, employing an AdamW optimizer\supercite{loshchilov_decoupled_2019}.
\subsection{SHAP value computation}
To approximate feature contributions for the outputs of our models, we calculated \ac{shap} values analogous to our previous work\supercite{ferle_predicting_2025}.
In brief, we used the DeepExplainer class\supercite{shrikumar_learning_2019, strumbelj_explaining_2014} of the \ac{shap} python library\supercite{lundberg_unified_2017} (version 0.46.0).
In a given cross-validation fold, the DeepExplainer was provided with the entire training partition as background datasets and the entire testing partition as the input (refer to methods section \textit{\nameref{sec:patient_characteristics}} for details about data partitioning).
The SHAP values were computed for each cross-validation fold individually and then combined in the final analysis.

        \section{Data availability}\label{sec:data_availability}
        The CoMMpass data is available upon registration in the \ac{mmrf} Researcher Gateway at \url{https://research.themmrf.org}.\\
The Lung1 data was originally part of a study by Aerts et al.\supercite{aerts_decoding_2014} and is publicly available at the \textit{Cancer Imaging Archive}\supercite{aerts_data_2019}.

        \section{Code availability}\label{sec:code_availability}
        We hosted a \href{https://github.com/maximilianferle/Unsupervised-risk-factor-identification-across-cancer-types-and-data-modalities-via-explainable-AI}{public GitHub repository} with all code used to produce the results of this study.

        \section{Acknowledgements}\label{sec:acknowledgements}
        The authors acknowledge the financial support by the Federal Ministry of Education and Research of Germany and by Sächsisches Staatsministerium für Wissenschaft, Kultur und Tourismus in the programme Center of Excellence for AI-research "Center for Scalable Data Analytics and Artificial Intelligence Dresden/Leipzig", project identification number: ScaDS.AI

This work was partially funded by grants from the International Myeloma Society (IMS Research Grant 2023), German Research Foundation (SPP µbone) and EU HORIZON (CERTAINTY).
The CERTAINTY project is funded by the European Union (Grant Agreement 101136379).
Views and opinions expressed are however those of the authors only and do not necessarily reflect those of the European Union or the Health and Digital Executive Agency.
Neither the European Union nor the granting authority can be held responsible for them.

The data used in this study were generated as part of the MMRF Personalized Medicine Initiatives (https://research.themmrf.org and https://themmrf.org).
We thank the MMRF for technical support and for facilitating access to their data.

The Lung1 data was originally part of a study by Aerts et al.\supercite{aerts_decoding_2014} and is publicly available at the \textit{Cancer Imaging Archive}\supercite{aerts_data_2019}.
We thank Aerts et al. for providing access to their data.

        \section{Author contributions}\label{sec:author_contributions}
        MF devised the main conceptual ideas, implemented the training algorithms, designed and implemented models, conducted experiments, analysed data, interpreted results, designed figures and took the lead in writing the manuscript.\\
JA processed and curated data, conducted experiments, interpreted results, designed figures and contributed to writing the manuscript.\\
TW, NG and AL contributed to conceptualizing the analyses.\\
HJM, TN, MK, KR and MM revised the manuscript contributed to interpreting results.\\
KR, and MM supervised the project and provided critical feedback on the presentation of the results.\\
All authors have read and approved the manuscript.

        \section{Competing interests}\label{sec:competing_interests}
        MM received financial support from
SpringWorks Therapeutics Inc.,
Janssen-Cilag GmbH and
Roche Deutschland Holding GmbH.\\
MF, JA, TW, NG, AL, HJM, TN, MK and KR declare no conflict of interest.

    \end{multicols}

    % Bibliography
    \printbibliography

%    \pagebreak
%    \input{sections/supplementals}

\end{document}